\documentclass[final]{cvpr}

\usepackage{times}
\usepackage{epsfig}
\usepackage{graphicx}
\usepackage{amsmath}
\usepackage{amssymb}

\usepackage{amstext}
\usepackage{amsopn}
\usepackage{booktabs}
\usepackage{multirow}
\usepackage{float}
\usepackage[numbers]{natbib}
\usepackage[american]{babel}
\usepackage{subfigure}

\newcommand{\yvec}{\mathbf{y}}
\newcommand{\xvec}{\mathbf{x}}
\newcommand{\Wmat}{\mathbf{W}}
\newcommand{\loss}{\mathcal{L}}

\def\tsc#1{\csdef{#1}{\textsc{\lowercase{#1}}\xspace}}
\tsc{WGM}
\tsc{QE}
\tsc{EP}
\tsc{PMS}
\tsc{BEC}
\tsc{DE}

\usepackage[pagebackref=true,breaklinks=true,colorlinks,bookmarks=false]{hyperref}

\begin{document}

\title{Adaptive Multi-Teacher Multi-level Knowledge Distillation}


\author {
  Yuang Liu\,,\quad Wei Zhang\thanks{Corresponding author.}\,,\quad Jun Wang \\
  East China Normal University, Shanghai, China \\
  {\tt\small \{frankliu624, zhangwei.thu2011, wongjun\}@gmail.com}
}

\maketitle


\begin{abstract}
Knowledge distillation~(KD) is an effective learning paradigm for improving the performance of lightweight student networks by utilizing additional supervision knowledge distilled from teacher networks. Most pioneering studies either learn from only a single teacher in their distillation learning methods, neglecting the potential that a student can learn from multiple teachers simultaneously, or simply treat each teacher to be equally important, unable to reveal the different importance of teachers for specific examples. To bridge this gap, we propose a novel adaptive multi-teacher multi-level knowledge distillation learning framework~(AMTML-KD), which consists two novel insights: (\romannumeral1) associating each teacher with a latent representation to adaptively learn instance-level teacher importance weights which are leveraged for acquiring integrated soft-targets~(high-level knowledge) and (\romannumeral2) enabling the intermediate-level hints~(intermediate-level knowledge) to be gathered from multiple teachers by the proposed multi-group hint strategy. As such, a student model can learn multi-level knowledge from multiple teachers through AMTML-KD. Extensive results on publicly available datasets demonstrate the proposed learning framework ensures student to achieve improved performance than strong competitors.
\end{abstract}

\section{Introduction}

Nowadays, with the broad prospect of artificial intelligence, more and more computational resources~(e.g., graphics processing units and large scale distributed systems) are invested in this area. It in turn leads to much larger statistical models, especially for deep neural networks~\cite{Hinton-NC2006}. From LeNet~\cite{LeCun-IEEE1998}, to AlexNet~\cite{Krizhevsky-NIPS2012}, and to VGG~\cite{Simonyan-ArXiv2014}, there is a clear trend that top-performing models are becoming larger with numerous parameters. Moreover, ensemble learning~\cite{Dietterich-IMCS2000} combines multiple models in an integrated framework, accentuating this trend and causing a much bigger size of model parameters. However, the heavy computational resources required by the approaches restrict their applications to platforms and devices with limited computational capability and low memory storage. Besides, many real application scenarios demand high speed execution, such as recommender systems~\cite{Bachrach-ACM2014}. Yet most complex models are slow to execute.

Recently two knowledge distillation learning paradigms, original KD~(OKD)~\cite{Hinton-ArXiv2015} and FitNet~\cite{Romero-ICLR2015} are proposed to transfer kno\-w\-ledge from a larger neural network~(teacher) to train a smaller and faster neural network~(student), while retaining high classification performance.
In original KD, the soft-target $\tilde{\yvec}^T$ generated by teacher is regarded as high-level knowledge.
It induces the following hybrid loss to let student mimic teacher output:
\begin{equation}
\loss_{OKD}= \mathcal{H}(\yvec, \yvec^S)+\lambda \mathcal{D}_{KL}(\tilde{\yvec}^{T}, \tilde{\yvec}^{S})\,,
\label{eq:eq1}
\end{equation}
where $\yvec$ and $\yvec^S$ correspond to the ground-truth and inferred labels of student, respectively.
$\tilde{\yvec}^S$ is the soft-target generated by student. $\mathcal{H}$ is the cross-entropy loss and $\mathcal{D}_{KL}$ is the Kullback Leibler~(KL) divergence. 
$\lambda$ is a hyper-parameter to control the relative influence of teacher knowledge transfer.

From another view, FitNet introduces intermediate-level knowledge from the teacher neural layers to guide the learning of student neural layers: 
\begin{equation}
\loss_{sHT}= \frac{1}{2}\lVert u(\xvec; \Wmat_{H}) - r\big(v(\xvec; \Wmat_{G});\Wmat_{r} \big) \rVert^2\,,
\label{eq:eq2}
\end{equation}
where $u$ and $v$ are the teacher/student deep nested functions up to their respective hint/guided layers with parameters $\Wmat_{H}$ and $\Wmat_{G}$. 
$r$ is a regression function with parameters $\Wmat_{r}$, which achieves the goal of training student to mimic the feature map of teacher. 
Since the teacher network is only involved in the training procedure and does not take up any resource for deployment and use, it is effective and efficient for applying the learned student model into practice.
It has aroused some successful applications in computer vision~\cite{Romero-ICLR2015,You-AAAI2018}, natural language processing~\cite{Hu-arXiv2016,Nakashole-NLP2017}, recommender system~\cite{Song-SIGIR2018}, etc.


Most prior studies of distillation learning go deeper into the direction of knowledge distillation from one teacher to one student.
However, there exist different well-performed teacher models with different complex architectures, or even with the same architecture but different learned parameters due to different initializations.  
We believe that learning from multiple teachers simultaneously has potential to fuse more useful knowledge and improve the performance of student.
Some recent studies~\cite{You-SIGKDD2017,FukudaSKTCR17} have taken a step along this direction by either simply treating different teachers equally or manually tuning the teacher importance weights, lacking the ability of automatically discriminating different teachers.
Some other studies~\cite{Ruder-ArXiv2017,Tan-ICLR2019,Vongkulbhisal-CVPR2019} have applied multiple teacher networks to multi-task or multi-domain settings. They assume each task or domain corresponds to a teacher network.
Therefore they are orthogonal to learning from multiple teacher networks for a single task in a single domain, which is a more general scenario used in most of knowledge distillation studies.


In this paper, we propose a novel Adaptive Multi-Teacher Multi-Level Knowledge Distillation learning framework, na\-med \textbf{AMTML-KD}, where the knowledge involves the high-level knowledge of soft-targets and the intermediate-level knowledge of hints from multiple teacher networks.
We argue that the fused knowledge is more comprehensive and effective for training a student network.
Specifically, for the high-level knowledge, we first associate a latent representation with each teacher to indicate its characteristics and adaptively determine the importance weights of different teachers with respect to a specific instance based on both the teacher representation and instance representation gotten from a student model.
These learned weights are leveraged for the weighted combination of soft-targets from corresponding te\-achers.
Regarding to the intermediate-level knowledge, we propose a multi-group hint strategy by letting the intermediate layers of each teacher in charge of a group of layers in the student network. It could be regarded as a extension of FitNet for learning from multiple teachers.
Through this manner, each teacher can transfer its deep feature representation to the intermediate layers of student. 

In summary, our main contributions are as follows:
\begin{itemize}
    \item We develop a novel distillation learning framework AMTML-KD, which is the first one to address the adaptive learning of multi-level knowledge from multiple teachers.

    \item We propose a computational method to determine the instance-level teacher importance weights for integrating soft-targets, and a multi-group hint strategy transfer intermediate-level knowledge. 

    \item We conduct extensive experiments on several public available datasets, verifying the effectiveness of our AMTML-KD, as well as the benefits brought by its main components.
\end{itemize}

\section{Related work}

Recent years have witnessed the large efforts applied to model compression, especially in the era of mobile computing.
~\cite{Bucilu-SIGKDD2006} is a pioneering study which utilizes trained large models to generate labels for unlabeled data, and then leverages the pseudo-labeled data as a complement to labeled training data to learn smaller models.
However, the knowledge of teachers lying in the training data is overlooked.
To address this issue, ~\cite{Ba-NIPS2014} utilizes labels generated by teachers for training data to optimize the student.
More recently, distillation learning paradigm~\cite{Hinton-ArXiv2015} is proposed to further combine both the ground-truth labels and teacher generated labels to guide the learning of a student model.
Note a similar learning strategy, privilege learning~\cite{Vapnik-JMLR2015}, focuses on using privilege information only existing in training stage to help optimizing models and is unified with distillation learning~\cite{Lopezpaz-ICLR2016}. 
Due to the strong applicability and generality of knowledge distillation, there are a great deal of follow-up studies.

In contrast to leveraging the generated soft-targets as kn\-owledge, some studies such as FitNet~\cite{Romero-ICLR2015} and attention transfer~\cite{Zagoruyko-ICLR2017} take the intermediate representations of teachers as another type of knowledge and guide student to mimic the representations.
Moreover, explicit knowledge from rules is incorporated into distillation learning through expectation regularization~\cite{Hu-arXiv2016,Song-SIGIR2018}.
More recently, relational knowledge distillation~(RKD)~\cite{Park-CVPR2019} considers a new form of knowledge called structural knowledge, which is derived from the teach\-er outputs of multiple examples.
This is different from previous studies which distills knowledge from each single example. 
Knowledge distillation also finds its wide range of applications.
Despite image classification, the tasks of visual question answering~\cite{Mun-NIPS2018}, semantic segmentation~\cite{Liu-CVPR2019} and machine translation~\cite{Tan-ICLR2019} have been also benefited from distillation learning.

Although rich forms of knowledge and various applications have been extensively investigated by prior studies, only a few studies have considered the distillation learning from another perspective, i.e., learning knowledge from multiple teachers.
~\cite{Sau-ArXiv2016} presents a noise-based regularization method, simulating learning from multiple teachers.
~\cite{Ruder-ArXiv2017} studies the domain adaption task by associating each source domain with a teacher for later distillation learning.
~\cite{Tan-ICLR2019} applies multi-teacher learning into multi-task learning where each teacher corresponds to a task.
Similarly,~\cite{Vongkulbhisal-CVPR2019} trains a classifier in each source and unifying their classifications on an integrated label space.
Nevertheless, the multi-domain or multi-task setting limits the application of the method to more general scenarios for a single task in a single domain.

The most relevant studies to this paper are as follows: ~\cite{FukudaSKTCR17} manually tunes the weights of each teacher, which is costly and not automatic. ~\cite{You-SIGKDD2017} equally fuses knowledge from multiple teachers in a more general setting. Unfortunately, since it assumes each teacher to be equally important and gets the integrated knowledge by averaging the soft-targets from different teacher networks, the different importance of different teachers for each data instance cannot be captured.
Two other studies~\cite{Zhang-CVPR2018,You-AAAI2018} train each student with knowledge from other students by the idea of basic distillation learning but without considering any teacher guidance, which are orthogonal to our investigation.

\section{Proposed learning framework}

In this section, we provide the motivation of learning teacher-dependent importance weight and an overview of the learning framework. Then we specify its mathematical details and give intuitive understandings.

\subsection{Motivation and overview}

Consider there are two teachers, in which one is a math teacher and the other is a psychology teacher. If there is a partial differential equation to be solved and the two teachers give some solutions, then how should students learn from these solutions? From the perspective of human cognition, students should learn more from the math teacher's solutions since they might contain more useful knowledge. The above intuition shows the necessity of considering different importance of teachers for different types of problems.

\begin{figure}
\centering
\includegraphics[width=1.0\columnwidth]{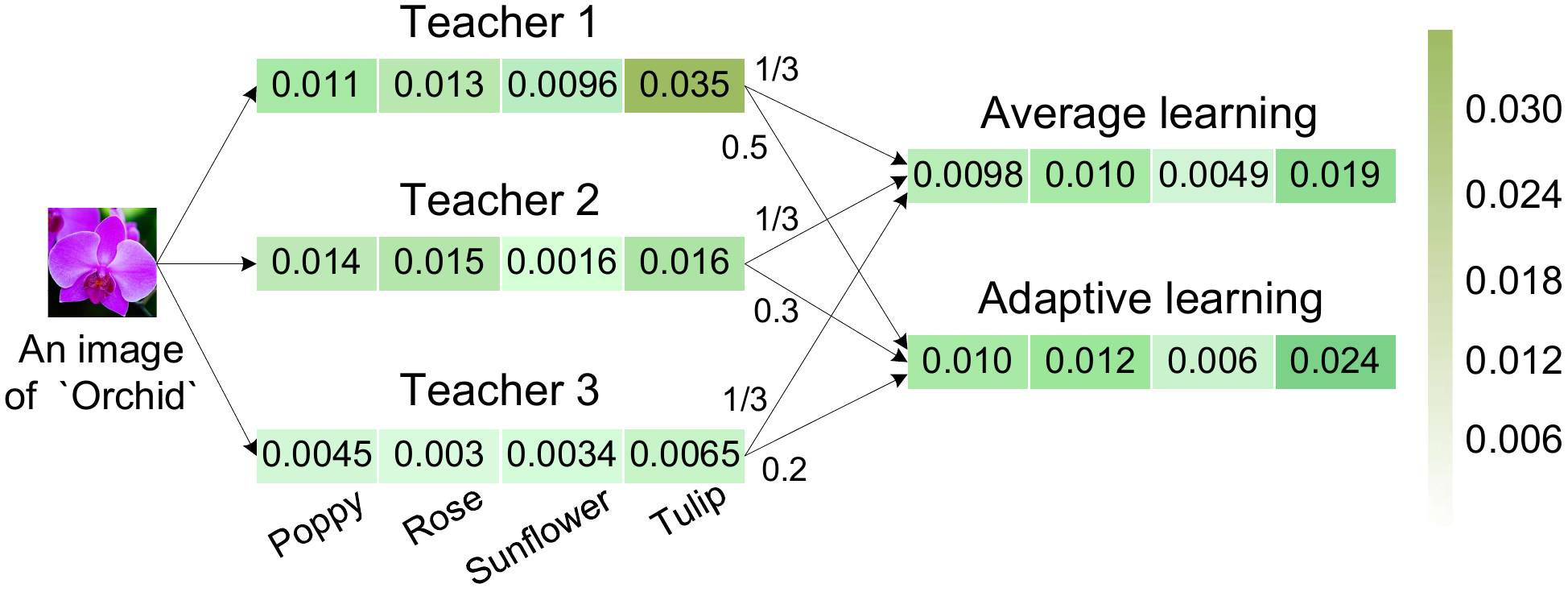} 
\caption{Illustration of adaptive learning using an example (orchid).
The corresponding soft targets are generated by three teacher models. Note: poppy, rose, sunflower, and tulip are the four categories we choose that are similar to orchid.}
\label{fig:fig1}
\end{figure}

Now we pay attention to the task of image classification which will be tested in the experiments. As the example shown in Figure \ref{fig:fig1}, given an image of orchid, three teacher neural networks have different prediction values for the same set of image categories. We could observe that the soft-target values generated from the first teacher carry more information of the similarities between the chosen labels and the ground truth label of which the prediction value is much larger and omitted in the figure. 
Therefore it is intuitive to allocate a larger weight to the first teacher for this image and smaller weights to the other two teachers to get the integrated soft-target value. Meanwhile, the deeper knowledge learned by teacher models can help student learn at each level.
The above discussion motivates the proposal of learning adaptively from multiple teachers and multiple levels.

\begin{figure*}
\centering
\includegraphics[width=0.85\textwidth]{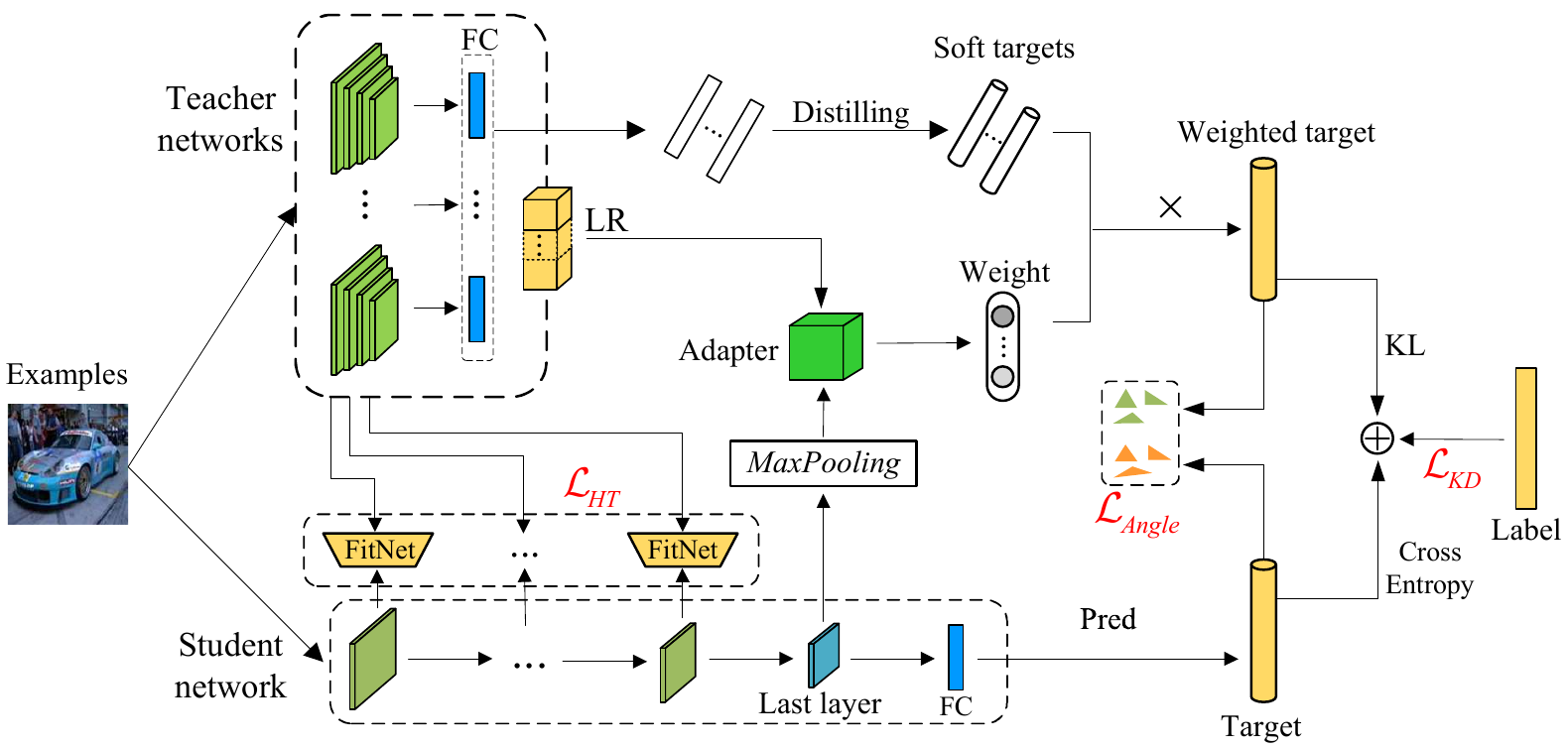}
\caption{The architecture of the adaptive multi-teacher multi-level knowledge distillation learning framework.}
\label{fig:fig2}
\end{figure*}

Figure \ref{fig:fig2} depicts the overview of the proposed adaptive learning framework. For each example in a given training set, we first employ several well-trained teacher networks (2 teachers shown in the figure) to generate the corresponding soft-target values as high-level knowledge and representations of intermediate layers as intermediate-level knowledge. 
The adapter is responsible for adaptively learning instance-level teacher importance weights for integrating soft-targets, which are further utilized to derive the standard knowledge distillation loss $\loss_{KD}$ and angle based loss $\loss_{Angle}$. $\loss_{KD}$ and  $\loss_{Angle}$ help to learn the weighted dark knowledge and structural relation between examples respectly. 
Moreover, a multi-group hint based loss $\loss_{HT}$ is for transferring intermediate-level knowledge from multiple teacher layers.
The overall optimization target of the proposed AMTML-KD is given as:
\begin{equation}
\mathcal{L}=\loss_{KD}+\alpha\mathcal{L}_{Angle} +\beta\loss_{HT}\,,
\end{equation}
where $\alpha$ and $\beta$ are hyper-parameters to control the relative influence of the corresponding losses in the objective.


\subsection{Adaptive learning of teacher importance }

To explicitly capture the implicit characteristics of teach\-er networks, we introduce a set of latent variables making up a latent representation(named LR in Figure \ref{fig:fig2}) to represent them. For example, the $t$-th teacher model is associated with a factor, denoted as $\boldsymbol{\theta}_{t} \in \mathbb{R}^{d}$ where d is the dimension of the factor and $t \in\{1, \cdots, m\}$ with $m$ teachers. This strategy is partially inspired by latent factor models commonly applied in the recommender system~\cite{Koren-SIGKDD2008}, where each user or item corresponds to one latent factor used to summarize their implicit features.

The representations of instances could be extracted from the output of some layer in a student network.
Since image is what we focus on in this work, we take the output of the last convolutional layer as the tensor representation of an image by convention~\cite{Nam-CVPR2017}.
As such, we get $\mathbf{B}_{i} \in \mathbb{R}^{C H W}$ for the $i$-th image where $C$, $H$ and $W$ correspond to the number of channel, height and width of student's feature map, respectively.
To convert the space of the image representation to be the same as that of the teacher factor for ease of later computation, we adopt a simple max-pooling operation with a kernel whose size $s=(H\times W)$.
It keeps the most significant value in each channel. The operation is represented as follows:
\begin{equation}
\boldsymbol{\delta}_{i}=\operatorname{MaxPooling}\left(\mathbf{B}_{i}, s\right)\,,
\end{equation}
where $\boldsymbol{\delta}_{i} \in \mathbb{R}^{C}$ and $d$ is set to be equal to $C$ for simplicity (e.g., 128).

Based on the above illustration, we calculate the importance weight of the $t$-th teacher model for the $i$-th image through the following manner: 
\begin{equation}
\gamma_{t, i}=\boldsymbol{\nu}^{\mathrm{T}}\left(\boldsymbol{\theta}_{t} \odot \boldsymbol{\delta}_{i}\right)\,,
\label{eq:eq5}
\end{equation}
where $\boldsymbol{\nu}$ is a global parameter vector to be learned, $\odot$ denotes element-wise product.
Larger $\gamma_{t, i}$ denotes the teacher is more important with regard to the image. From Equation \ref{eq:eq5}, we can observe the interaction between the representations of the teacher model and the image is captured by the element-wise product operation.
It computes their similarities in each dimension. $\boldsymbol{\nu}$ determines whether or not the value in each dimension has a positive effect on the score. We further normalize the importance weight through the $\operatorname{softmax}$ function defined as follows:
\begin{equation}
{w}_{t, i}=\operatorname{softmax}\left(\gamma_{t, i}\right)=\frac{\exp \left(\gamma_{t, i}\right)}{\sum_{t^{\prime}=1}^{m} \exp \left(\gamma_{t^{\prime}, i}\right)}\,.
\end{equation}

In contrast to the average learning strategy~\cite{You-SIGKDD2017}, we propose the weighted addition operation to acquire the integrated soft-target $\tilde{\yvec}_{i}^{T}$, which is defined as below:
\begin{equation}
\tilde{\yvec}_{i}^{T}=\sum_{t=1}^{m} {w}_{t, i} \tilde{\yvec}_{t,i}^T\,,
\label{eq:eq7}
\end{equation}
where $\tilde{\yvec}_{t,i}^T$ is the soft-target generated by the $t$-th teacher for the $i$-th image.

\subsection{Learning high-level knowledge }

Given the high-level knowledge revealed in integrated soft-targets, we adopt two ways of utilizing the knowledge to guide the learning of student. 

\subsubsection{Standard Knowledge Distillation}

Following the standard knowledge distillation loss shown in Equation \ref{eq:eq1}, we define our objective by the incorporation of fused multi-teacher knowledge, which is given as:
\begin{equation}
\loss_{KD}=\sum_{i} \mathcal{H}\left(\yvec_i, \yvec_i^S\right)+
\lambda \sum_{i}\mathcal{D}_{KL}\left(\tilde{\yvec}_{i}^{T}, \tilde{\yvec}_i^{S}\right)\,,
\end{equation}
where $\tilde{\yvec}_i^{S}$ is the soft-target of the $i$-th image generated by the student network and $\tilde{\yvec}_i^{T}$ is the integrated soft-target computed by Equation~\ref{eq:eq7}.

\subsubsection{Structural knowledge distillation}

In addition to the standard knowledge distillation which regards each data example separately, we also consider to add structural knowledge between different data examples to transfer relational information effectively, which is motivated by~\cite{Park-CVPR2019}.
Concretely, given a triplet of examples denoted as $(i, j, k)$, an angle based metric is used to measure the structural relation between these examples, which is given as:
\begin{equation}
\Delta\left(x_{i}, x_{j}, x_{k}\right)=\cos \angle x_{i} x_{j} x_{k}=\left\langle\mathbf{e}^{i j}, \mathbf{e}^{k j}\right\rangle\,,
\end{equation}
where $\mathbf{e}^{i j}$ and $\mathbf{e}^{k j}$ are normalized vector differences, i.e., $\mathbf{e}^{i j}=\frac{x_{i}-x_{j}}{\left\|x_{i}-x_{j}\right\|_{2}}, \mathbf{e}^{k j}=\frac{x_{k}-x_{j}}{\left\|x_{k}-x_{j}\right\|_{2}}$. 



Based on the obtained integrated soft-targets, we can define our angle based loss function under the scenario of multiple teachers as follows:
\begin{equation}
\loss_{Angle}=\sum_{\left(i, j, k\right)} l_{\delta}\big(\Delta\left(\tilde{\yvec}_{i}^{T}, \tilde{\yvec}_{j}^{T}, \tilde{\yvec}_{k}^{T}\right), \Delta\left(\tilde{\yvec}_{i}^{S}, \tilde{\yvec}_{j}^{S}, \tilde{\yvec}_{k}^{S}\right)\big)\,,
\end{equation}
where $l_{\delta}$ is the Huber loss which provides robust regression, compared with standard mean square loss. Note that $\tilde{\yvec}_i^{T}$, $\tilde{\yvec}_j^{T}$ and $\tilde{\yvec}_k^{T}$ are all integrated soft-targets computed by Equation~\ref{eq:eq7} for different image examples.

\subsection{Learning intermediate-level knowledge}

Deep neural networks enjoy the advantage of feature representation learning in their intermediate layers and thus it is promising to transfer this intermediate-level knowledge to student.
Motivated by FitNet~\cite{Romero-ICLR2015} which learns intermediate-level knowledge from a single teacher, we propose a simple but effective extension of FitNet to enable the learning from multiple teachers, named multi-group hint.
In particular, the method makes each teacher to be responsible for a group of layers of the student network.
Through experiments, we empirically found that student could benefit from the representation of deep layer learned by teacher.
As such, we utilize the last feature layer of each teacher to guide knowledge distillation and define a multi-group hint based loss as follows:
\begin{equation}
\loss_{HT}= \sum_{t, l=f(g)}^{t=m}\lVert u_{t} - \mathbf{F}_t(v_{l})\rVert^2\,,
\end{equation}
where $u_{t}$ is the last layer feature map of the teacher $t$, $v_{l}$ corresponds to the feature map output by the $l$-th group layers in student. $\mathbf{F_t}$ is a single layer FitNet for teacher $t$ to make the feature maps adaptive. Note that $f(\cdot)$ specifies the one-to-one mapping from a group of student layers to a teacher, and it can be chosen flexibly.

We propose several mapping strategies: (\romannumeral1) assigning tea\-chers with better performance to high-layer groups; and (\romannumeral2) assigning teachers with better performance to low-layer gro\-ups; and (\romannumeral3) random assignment of teachers to different gro\-ups.
The results on different datasets have consistently indicated that the first strategy gain a little improvement over the other two strategies.
We therefore choose it as our default mapping strategy and report its results in the experiments.

\begin{table*}
\centering
\caption{Student and teacher models trained on CIFAR-10 and CIFAR-100.}\label{tab:tab1}
\setlength{\tabcolsep}{6mm}{
\begin{tabular}{c|c|c|cc}
\toprule
Models & \#Params & Compression rate & CIFAR-10 & CIFAR-100\\\midrule
Stu1 & $\sim1.13\text{M}$ & $\times138.05$ & 92.29\% & 68.69\%\\
Stu2 & $\sim295\text{K}$ & $\times528.81$ & 88.71\% & 58.50\%\\
Stu3 & $\sim595\text{K}$ & $\times262.18$ & 90.48\% & 62.02\%\\\hline
ResNet110 & $\sim7.06\text{M}$ & $\times22.10$ & 93.87\% & 73.45\%\\
VGG-19 & $\sim156\text{M}$ & $\times1$ & 93.51\% & 73.40\%\\
DenseNet121 & $\sim28.3\text{M}$ & $\times5.51$ & 95.41\% & 79.74\%\\\bottomrule
\end{tabular}
}
\end{table*}

\section{Experimental setup}

In this section, we introduce the datasets used in the experiments, the baselines we considered, and the implementation details.




\subsection{Datasets}

\subsubsection{CIFAR-10 \& CIFAR-100} 
Both CIFAR-10 and CIFAR-100 datasets consist of the same 50000 training images and 10000 testing images.
The major difference between the two datasets is that CIFAR-10 involves 10 equal-sized categories while CIFAR-100 consists of 100 categories, which means image classification on CIFAR-100 is more challenging.

\subsubsection{Tiny-ImageNet}
The dataset covers 200 image classes from ImageNet~\cite{Russakovsky-IJCV2015} with 500 training examples per class.
It also has 50 validation and 50 test examples per class.
All the images in the dataset are down-sampled to 64x64 pixels while their original size is 256x256.
Some basic data augmentation operations are adopted on all three datasets, such as feature-wise centralization and normalization, horizontal flipping, etc.

\subsection{Baselines}
We compare against the following representative baselines, including standard and state-of-the-art knowledge distillation models for a single teacher and multiple teacher bas\-ed knowledge distillation method, as well as deep mutual learning with multiple students:

\textbf{OKD~\cite{Hinton-ArXiv2015}:} This is the pioneering work of introducing the original knowledge distillation~(OKD) learning for one teacher and one student networks.
A naive extension of OKD to learn from multiple teacher is to traverse all teachers and each time select a teacher network to teach student.

\textbf{FitNet~\cite{Romero-ICLR2015}:} FitNet is the standard approach to utilize intermediate-level hints from hidden layers of one teacher to guide the training of student layers.

\textbf{RKD~\cite{Park-CVPR2019}:} Relational knowledge distillation~(RKD) is the state-of-the art KD method, aims at transferring structural knowledge, accompanied by intermediate-knowledge transfer~\cite{Yim-CVPR2017} which limits the architecture or intermediate layer shape must be consistent.

\textbf{AvgMKD~\cite{You-AAAI2018}}: This approach distills knowledge from multiple teachers by: 1) treating each teacher equally when calculating integrated soft-targets; and 2) transferring relative dissimilarity~(RD) relationships among the intermediate representations of teacher networks.

\textbf{DML~\cite{Zhang-CVPR2018}}: Deep mutual learning~(DML) utilizes the basic distillation learning from the perspective of training multiple student networks.

For ease of illustration, we use OKD(X) to denote the adoption of teacher X, which is the same for FitNet(X) and RKD(X).
DML($n$S) denotes using $n$ students in mutual leanring.
AvgMKD($m$T) indicates that AvgMKD leverages $m$ teacher models, and this principle of notation also applies to other multiple teacher based approaches, including our proposed AMTML-KD.

\subsection{Implementation Details}
We tuned the hyper-parameters of all adopted methods based on the performance on the validation datasets for fair comparison.
Specifically, the temperature for computing soft-targets is set to 5.0 and $\lambda=0.7$ in knowledge distillation loss.
We set $\alpha=1$ and $\beta=2$ in our optimization target by default.
SGD with a mini-batch size of 128 is adopted to optimize our model.
The learning rate is initialized by 0.1 and decayed at in 100 and 150 epochs.
We run all models on a single GPU~(GTX 1080Ti). 



\begin{table}
\centering
\caption{Results of various KD methods for Stu1 on CIFAR-10 and CIFAR-100.}\label{tab:tab2}
\begin{tabular}{c|cc}
\toprule
Methods& CIFAR-10& CIFAR-100\\\midrule
Stu1& 92.29\% & 68.69\%\\\hline
OKD(ResNet110)& 92.62\% & 69.05\%\\
OKD(VGG-19)& 92.57\% & 68.92\%\\
OKD(DenseNet121)& 92.59\% & 68.98\%\\
FitNet(DenseNet121)& 92.88\% & 69.26\%\\
RKD(ResNet110)& 92.91\% & 69.94\%\\\hline
DML(3S)& 92.70\% & 68.99\%\\
AvgMKD(3T)& 93.01\% & 69.64\%\\\hline
\textbf{AMTML-KD(3T)}& \textbf{93.64\%} & \textbf{70.39\%}\\\bottomrule
\end{tabular}
\end{table}

\begin{table*}
\centering
\caption{Results of multi-teacher KD methods for three students on CIFAR-10 and CIFAR-100.}\label{tab:tab3}
\setlength{\tabcolsep}{4mm}{
\begin{tabular}{c|ccc|ccc} 
\toprule
\multirow{2}*{Methods}& \multicolumn{3}{c}{CIFAR-10}& \multicolumn{3}{|c}{CIFAR-100}\\\cline{2-7}
 & Stu1 & Stu2 & Stu3 & Stu1 & Stu2 & Stu3\\\midrule
DML(3S) & 92.70\% & 88.94\% & 90.43\% & 68.99\% & 58.87\% & 62.46\%\\
AvgMKD(3T) & 93.01\% & 89.32\% & 90.83\% & 69.64\% & 59.21\% & 62.71\%\\
\textbf{AMTML-KD(3T)} & \textbf{93.64\%} & \textbf{89.89\%} & \textbf{91.47\%} & \textbf{70.39\%} & \textbf{60.06\%} & \textbf{63.36\%}\\\bottomrule
\end{tabular}
}
\end{table*}

\begin{table*}
\centering
\caption{Performance of different students learned by AMTML-KD on CIFAR-10.}\label{tab:tab4}
\setlength{\tabcolsep}{6mm}{
\begin{tabular}{c|ccc}
\toprule
\#Teachers & Stu1 & Stu2 & Stu3\\\midrule
2T\{ResNet110/56\} & 93.20\% & 89.33\% & 90.98\%\\
2T\{VGG-19,DenseNet121\} & 93.11\% & 89.70\% & 91.09\%\\\hline
3T\{ResNet110/56/32\} & 93.46\% & 89.77\% & 91.48\%\\
3T\{ResNet110,VGG-19,DenseNet121\} & 93.64\% & 89.89\% & 91.47\%\\\hline
5T\{ResNet110/56/32,VGG-19,DenseNet121\} & 93.69\% & 90.02\% & 91.52\%\\
\bottomrule
\end{tabular}
}
\end{table*}

\section{Experimental results}

We verify the effectiveness of our proposed  AMTML-KD through the experimental results on  CIFAR-10, CIFAR-100, and Tiny-ImageNet.
Since the CIFAR-10 and CIFAR-100 have the same datasets except different number of labels, we first present the experiments on them together.
It is followed by the experiments on Tiny-ImageNet.
In the end, we perform ablation study to show the contribution of each component to the final model. 
For evaluation measure, we adopt the top-1 image classification accuracy.

\subsection{Results on CIFAR}

\subsubsection{Comparison with baselines} With respect to the two CIFAR datasets, we adopt three teacher neural networks, i.e., ResNet110, VGG-19, and Den\-seNet121.
We regard ResNet20 as the default student network~(Stu1 for short) and design another two small residual networks as the second and third students.
Table~\ref{tab:tab1} shows the number of model parameters and their performance on CIFAR-10 and CIFAR-100, from which we can see the chosen student networks are with much smaller size than the teacher models and their results are no doubt worse.
We denote the compression rate from the perspective of VGG-19 since it is the largest teacher here. 
The highest compression rate is 528.81 by the Stu2 model, with only 295KB of parameters.

Table~\ref{tab:tab2} shows the results on the two datasets by our approach and compared methods.
The student model trained alone perform not as well as all the other knowledge distillation methods, showing the benefit of distilling knowledge from teacher networks.
The models in the second part of the table are a group single-teacher distillation models. 
By performing comparison within this group, we can observe FitNet behaves better than standard knowledge distillation based methods and the recent RKD model achieves the state-of-the-art performance.
The third part of the table includes a multiple student model and a multiple teacher model.
Although DML(3S) has three student networks to learn collaboratively with each other, the performance of its best student model is worse than AvgMKD(3T).
This is due to the limitation of learning from others under training.
Later we show that a direct incorporation of multiple teachers into DML can indeed improve results. 
Finally, we can see that our AMTML-KD(3T) model outperforms all the other models by improving the state-of-the-art AvgMKD(3T) model on CIFAR-10 from 93.01\% to 93.64\%.
It implies that our approach could actually incorporate the informative knowledge from multiple teachers more effectively.

It is worth noting that AMTML-KD(3T) retains nearly the same number of parameters as student has since the new parameters added are only $\boldsymbol{\nu}$ and $\boldsymbol{\theta}_{t}(t \in (1,2,3))$ whose numbers are very small and could be ignored. 



\subsubsection{Performance on different student networks} 
We further analyze the effectiveness of our learning met\-h\-od for different student networks. Table~\ref{tab:tab3} shows the corresponding results.Although different students do not perform equally well, our AMTML-KD approach consistently outperforms AvgMKD and DML, showing its potential wide application.

\begin{figure}
\centering
\includegraphics[width=1.0\linewidth]{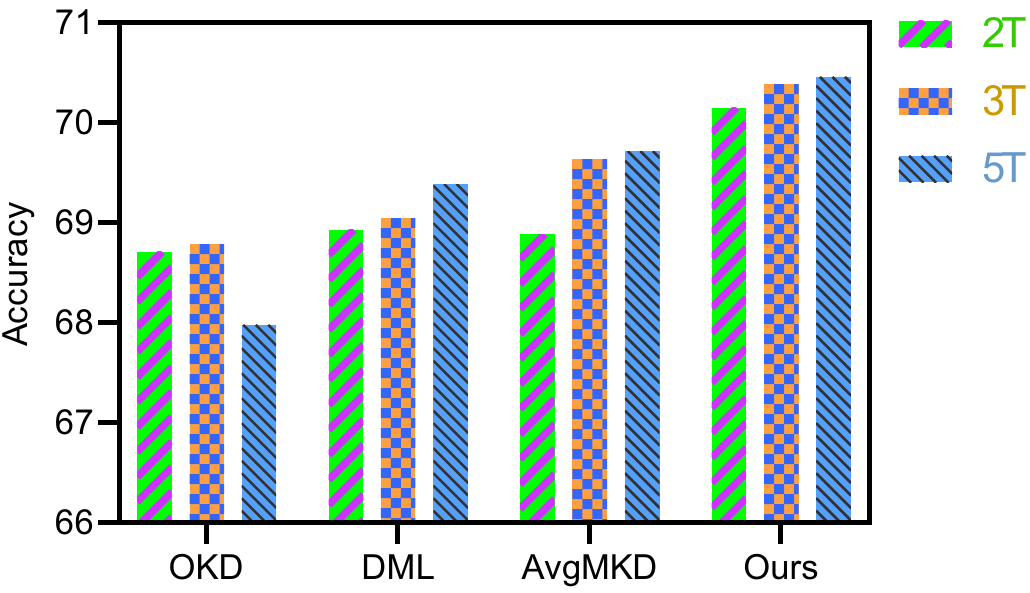}
\caption{Comparison of multiple-teacher KD methods with different number of teachers on CIFAR-100.
Note that in this experiment, DML is simply extended to a multiple teacher version DML(3T), by regarding teachers as students and mutually learns them with a student network.}
\label{fig:fig3}
\end{figure}

\subsubsection{Effect of the number of teachers}
As complementary experiments, we investigate how the number of teacher models affects the performance of our learning approach.
We first show whether the effect is consistent across different student networks.
Table~\ref{tab:tab4} shows the results corresponding to the first aspect, from which we can observe a clear trend that as the number of teachers increases, all the student networks achieve better results.
Besides, the performance gap between 2 teachers and 3 teachers is larger than the gap between 3 teachers and 5 teachers, showing the performance gain becomes smaller when the number of teachers continues to grow.
We further show the comparison of multiple teacher based knowledge distillation methods when having different number of teachers.
The results revealed in Figure~\ref{fig:fig3} 
verify that our approach is consistently better as well.

\begin{table}[th]
  \centering
  \caption{Comparison of GPU memory and training time cost on ResNet20.}
  \label{tab:tab9}
  \begin{tabular}{ccc}
  \toprule
  Methods& GPU memory& Training time\\
  \midrule
  individual & 0.861G & 32.4min \\
  OKD & 0.907G &42.5min \\
  FitNet & 0.921G & 44.2min \\
  \hline
  DML(3S) & 3.195G & 95.6min \\
  AvgMKD(3T) & 1.332G & 73.4min \\
  AMTML-KD(3T) & 1.351G & 75.8min\\
  \bottomrule
  \end{tabular}
\end{table}
    
Table~\ref{tab:tab9} shows the comparison of GPU memory and training time cost when training ResNet20~(Stu1) on CIFAR-100. The single teacher KD methods which usually have uncompetitive performance cost less training time and GPU memory. But in multi-teacher methods, ours only need a little more training time and memory, due to the extra parameters from FitNets and adapter which even can be ignored compared with the large models. Since DML needs to update the parameters of all student networks at the same time, it will consume more resources. 

\subsection{Results on Tiny-ImageNet}
For Tiny-ImageNet, we construct three ResNet based neural networks as our teachers, and choose the same three student networks used in the CIFAR datasets as students.
The model parameter size and performance of the teachers, as well as the first student are shown in Table~\ref{tab:tab5}.

\begin{table}[h]
\centering
\caption{Student and teacher models on Tiny-ImageNet.}\label{tab:tab5}
\begin{tabular}{c|c|c}
\toprule
Models & \#Params & Tiny-ImageNet\\\midrule
Stu1 & $\sim1.13\text{M}$ & 54.10\%\\\hline
ResNet110 & $\sim7.06\text{M}$ & 55.62\%\\
ResNet56 & $\sim3.49\text{M}$ & 54.65\%\\
ResNet32 & $\sim1.92\text{M}$ & 54.40\%\\\bottomrule
\end{tabular}
\end{table}

Table~\ref{tab:tab6} compares our approach with the baselines on the Tiny-ImageNet dataset, taking the first student as an example.
Similar conclusions can be drawn: (1) RKD achieves the state-of-the-art performance in single-teacher models; (2) the student model guided by multiple teachers obtains better results than single-teacher models, although AvgMKD\\(3T) behaves nearly the same as RKD; (3) the proposed meth\-od AMTML-KD achieves the best result, showing its performance is robust for different datasets.

\begin{table}[h]
\centering
\caption{Results of KD methods for Stu1 on Tiny-ImageNet.}\label{tab:tab6}
\begin{tabular}{c|c}
\toprule
Methods & Tiny-ImageNet\\\midrule
Stu1 & 54.10\%\\\hline
OKD(ResNet110) & 54.61\%\\
OKD(ResNe56) & 54.36\%\\
OKD(ResNet32) & 54.23\%\\
FitNet(ResNet110) & 54.85\%\\
RKD(ResNet110) & 54.94\%\\\hline
DML(3S) & 54.78\%\\
AvgMKD(3T) & 55.03\%\\\hline
\textbf{AMTML-KD(3T)} & \textbf{55.67\%}\\\bottomrule
\end{tabular}
\end{table}

\begin{table}[htb]
  \centering
  \caption{Results of multi-teacher KD methods for three students on Tiny-ImageNet.}\label{tab:tab7}
  \begin{tabular}{c|ccc}
  \toprule
  \multirow{2}*{Methods}& \multicolumn{3}{|c}{Tiny-ImageNet}\\\cline{2-4}
   & Stu1 & Stu2 & Stu3\\\midrule
  DML(3S) & 54.78\% & 50.86\% & 53.02\%\\
  AvgMKD(3T) & 55.03\% & 51.17\% & 53.38\%\\
  \textbf{AMTML-KD(3T)} & \textbf{55.67\%} & \textbf{52.08\%} & \textbf{53.99\%}\\\bottomrule
  \end{tabular}
\end{table}

Table~\ref{tab:tab7} presents the performance of our learning approa\-ch and the two baselines adopted on three different student networks.
We can find AMTML-KD indeed boosts performance of the three students over AvgMKD, which is consistent with the results on the CIFAR datasets.  


\subsection{Ablation study}
In this part, we conduct ablation study to verify the effectiveness of the main components in our approach AMTML-KD.
Table~\ref{tab:tab8} shows the performance of some baselines and variants of AMTML-KD, where the first part corresponds to the baselines and the second part denotes the variants.
To be specific, ``w/o $\loss_{Angle}\&\loss_{HT}$'' denotes removing the angle based loss and multip-group hint loss from AMTML-KD.
Analogously, ``w/o $\loss_{Angle} (\alpha=0)$'' and ``w/o $\loss_{HT} (\beta=0)$'' correspond to not considering $\loss_{Angle}$ and $\loss_{HT}$, respectively. 
``AvgMKD(3T) w/o RD'' denotes the removal of its RD loss, measuring relative dissimilarity relationships among intermediate representations of networks.

We first compare ``w/o $\loss_{Angle}\&\loss_{HT}$'' with ``AvgMKD\\*(3T) w/o RD'' and OKD(ResNet110), all of which use standard knowledge distillation objective but have different soft-targets.
The following observations are drawn from the comparison: 1) average learning from multiple teacher networks seems a little questionable for its not satisfied performance; and 2) adaptive learning from multiple teachers has its intrinsic benefits.
The comparison between FitNet and multi-group hint, the extension of FitNet proposed in this paper, shows the latter could bring benefits to some extent.

\begin{table}[htbp]
\centering
\caption{Ablation study of AMTML-KD.}\label{tab:tab8}
\scalebox{1}{
\begin{tabular}{c|cc}
\toprule
Methods & CIFAR-100 & Tiny-ImageNet\\\midrule
OKD(ResNet110)& 69.05\% & 54.61\%\\
AvgMKD(3T) w/o RD & 66.45\% & 53.03\%\\
FitNet & 69.26\% & 54.85\% \\\hline
Multi-group hint & 69.53\% & 55.00\%\\
w/o $\loss_{Angle}\&\loss_{HT}$ & 69.68\% & 54.92\%\\
w/o $\loss_{Angle} (\alpha=0)$  & 69.96\% & 55.43\%\\
w/o $\loss_{HT} (\beta=0)$ & 69.79\% & 55.21\%\\\hline
AMTML-KD(3T) & 70.39\% & 55.67\%\\
\bottomrule
\end{tabular}
}
\end{table}

The third part of Table~\ref{tab:tab8} shows the results of removing different loss functions from our approach.
Finally, we compare our full model with its variants ``w/o $\loss_{Angle} (\alpha=0)$'' and ``w/o $\loss_{HT} (\beta=0)$'', which shows that our improvements are not only from standard knowledge distillation with fused multi-teacher teacher, but also from the learning of structural knowledge and intermediate-level knowledge.  Although we use the existing RKD method, we only consider it to be more effective in combination with our adaptive method. If it is removed, the accuracy will be reduced by about 0.2\%. In summary, adapter determines the importance of each teacher which make knowledge distillation more flexible, HT loss helps improve the training of student through intermediate layer and adaptive angle loss could transfer relation information of examples to student.

\begin{figure}[htb]
  \centering
  \subfigure[Example `plane' on CIFAR-10]{
  \includegraphics[width=0.99\linewidth]{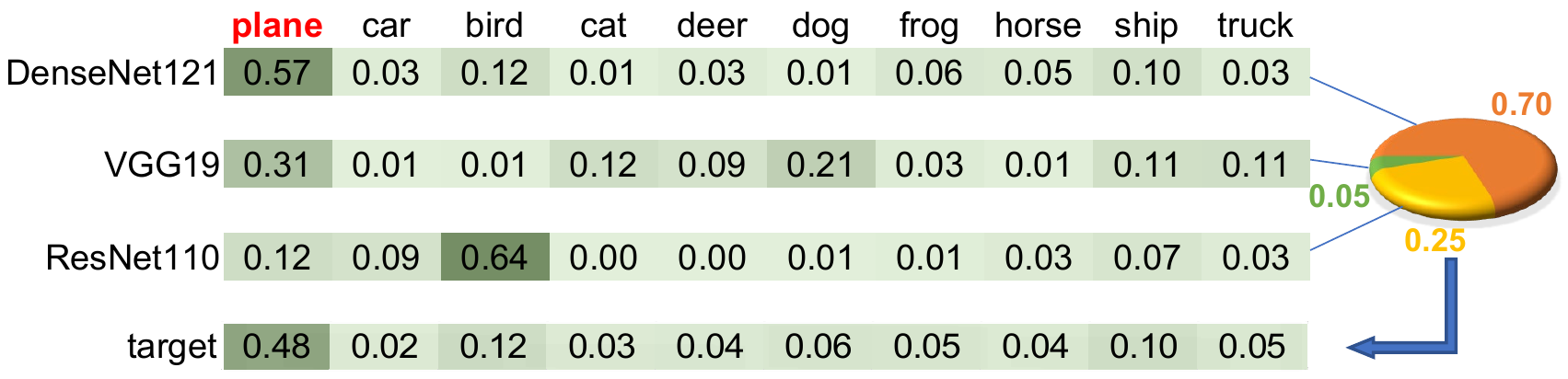}
  \label{subfig_a}
  }
  \quad
  \subfigure[Example `dog' on CIFAR-10]{
  \includegraphics[width=0.99\linewidth]{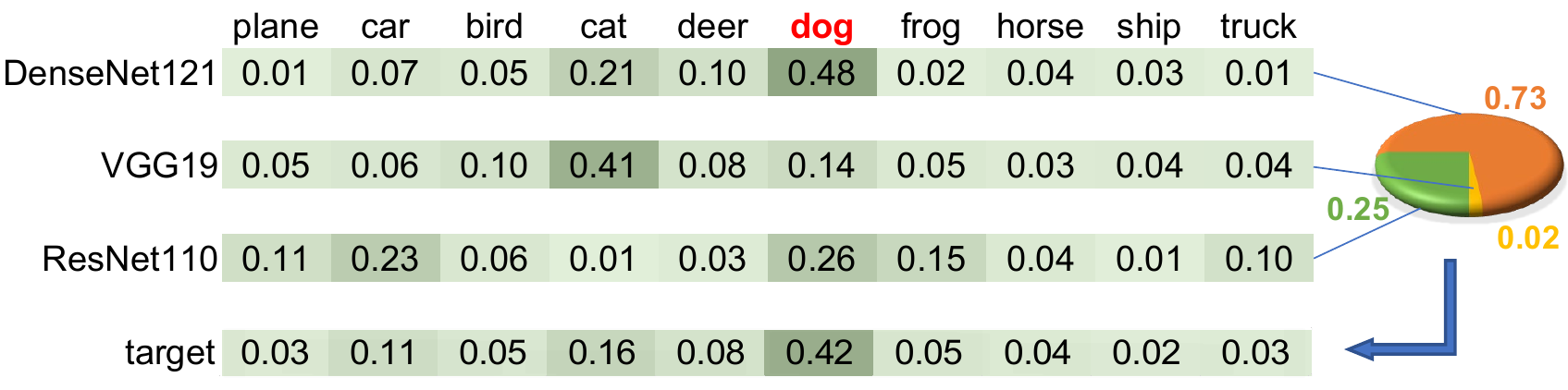}
  \label{subfig_b}
  }
  \quad
  \subfigure[Example `2' on MNIST]{
  \includegraphics[width=0.99\linewidth]{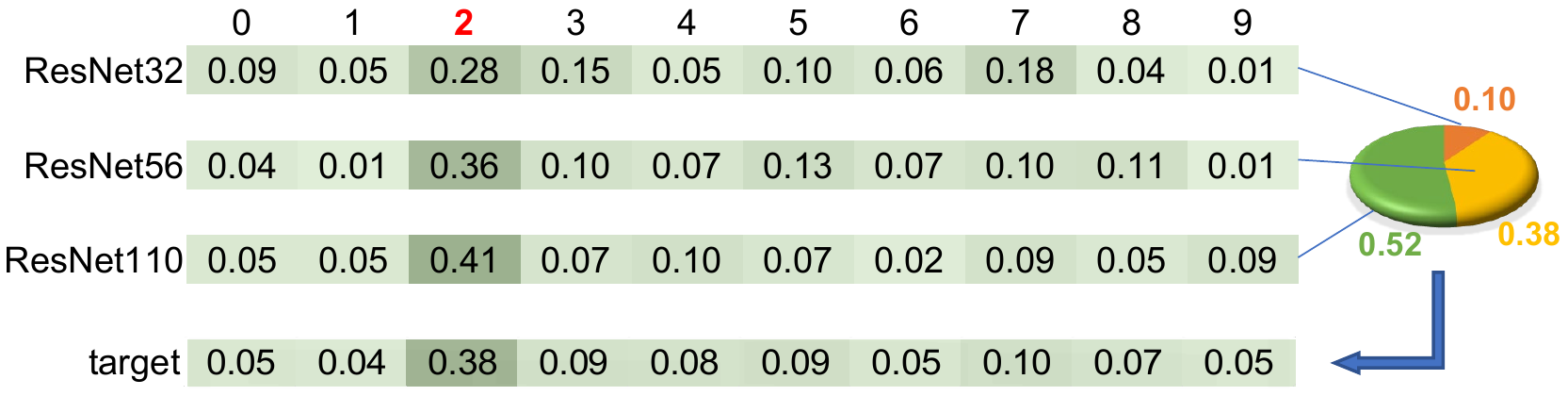}
  \label{subfig_c}
  }
  \caption{Visualization results of multi-teacher weights.}
  \label{fig:vis_weight}
\end{figure}

To verify importance of different weights, we show the visualization results of different weights for some examples on CIFAR-10 and MNIST in Figure~\ref{fig:vis_weight}. The bold red labels are the ground truth of the examples. We can see that even if some teachers' predictions are wrong, the adverse effects of these mistakes can be shielded by weight allocation, such as shown in Figure~\ref{subfig_a} and~\ref{subfig_b}.

\section{Conclusion}

In this paper, we have developed adaptive multi-teacher multi-level knowledge distillation~(AMTML-KD) learning framework. 
AMTML-KD learns distinct importance weights for different teacher networks w.r.t a specific data instance, ensuring better integration of soft-targets from multiple teac\-hers for transferring high-level knowledge.We have also proposed a simple multi-group hint strategy to enable AMTML-KD to learn intermediate-level knowledge from multiple tea\-chers.
Experimental results on image classification have de\-m\-onstrated AMTML-KD achieves the state-of-the-art performance and verified the effectiveness of its key components.

\section*{Acknowledgment}
This work was supported in part by NSFC (61702190, 61672236, 61672231), and NSFC-Zhejiang (U1609220).

{\small
\bibliographystyle{ieee_fullname}
\bibliography{ref}
}

\end{document}